\def\year{2021}\relax
\documentclass[letterpaper]{article}
\usepackage{aaai21}
\usepackage{times}
\usepackage{helvet}
\usepackage{courier}
\usepackage{graphicx}
\usepackage{natbib}  % DO NOT CHANGE THIS AND DO NOT ADD ANY OPTIONS TO IT
 \usepackage{booktabs}
 \usepackage[table,xcdraw]{xcolor}

\title{Trusting Machine Learning Results from Medical Procedures in the Operating Room}
\author {Ali El-Merhi\textsuperscript{\rm 1,2},  
Helena Odenstedt Herg{\'e}s\textsuperscript{1,2},
        Linda Block\textsuperscript{\rm 1,2}, 
        Mikael Elam\textsuperscript{\rm 3}, 
        Richard Vithal\textsuperscript{\rm 1,2}, 
        Jaquette Liljencrantz\textsuperscript{\rm 3},
        Miroslaw Staron\textsuperscript{\rm 4} \\
}

\affiliations{
    \textsuperscript{\rm 1}Institute of Clinical Sciences, Sahlgrenska Academy, University of Gothenburg, Gothenburg, Sweden, \\ 
    \textsuperscript{\rm 2} Department of Anesthesia and Intensive Care, Sahlgrenska University Hospital, Gothenburg, Sweden, \\
    \textsuperscript{\rm 3} Institute of Neuroscience and Physiology, Sahlgrenska Academy, University of Gothenburg, Gothenburg, Sweden \\
    \textsuperscript{\rm 4} Computer Science and Engineering, Chalmers $|$ University of Gothenburg, Gothenburg, Sweden
}

\begin{document}
\maketitle
% https://taih21.github.io/

\begin{abstract}
\begin{quote}
Machine learning can be used to analyse physiological data for several purposes. Detection of cerebral ischemia is an achievement that would have high impact on patient care. We attempted to study if collection of continous physiological data from non-invasive monitors, and analysis with machine learning could detect cerebral ischemia in tho different setting, during surgery for carotid endarterectomy and during endovascular thrombectomy in acute stroke. We compare the results from the two different group and one patient from each group in details. While results from CEA-patients are consistent, those from thrombectomy patients are not and frequently contain extreme values such as 1.0 in accuracy. We conlcude that this is a result of short duration of the procedure and abundance of data with bad quality resulting in small data sets. These results can therefore not be trusted.
\end{quote}
\end{abstract}

\section{Introduction}
\label{sec:introduction}
\noindent
Collecting patient data and using it for medical diagnoses by applying artificial intelligence has gained a lot of popularity in the last decade \cite{komorowski2019artificial}. Machine learning algorithms can identify patterns in data which are, otherwise, difficult to observe by a physician \cite{carra2020data}. An example of such situation is using artificial intelligence to detect cerebral ischemia, brain injury due to low blood flow, based on changes in heart rate variability \cite{odenstedt2021machine}. Several databases have been developed to provide the community with the reusable data sets, in turn providing better generalizability of results \cite{saeed2011multiparameter}. 

The performance of the machine learning model is based on the quality of the data collected. In the case of measurements performed outside of the operating room, there is usually enough time to set up sensors and check the quality of the signals before collecting data. However, in some cases, there is not enough time and therefore the quality of the signals can vary -- one of such procedures is thrombectomy, where a blood clot is removed from patient's brain. 

In this paper, we explore the differences in results from two procedures -- carotid endarterectomy (CEA), which allows for data quality control before data collection and endovascular thrombectomy ("thrombectomy"), which allows only for set-up due to the limited time. Our data collection methods are the same; we use ECG, EEG, NIRS, and RESP signals, which we quantify, featurize and use as input to machine learning algorithms CART, RandomForest and AdaBoost, support vector machines. The data was collected as part of a research project aiming to study if cerebral ischemia can be detected in real time by analysing data from bedside non-invasive monitors \cite{block2020cerebral}.

Our results show that the results from the thrombectomy cases vary in terms of accuracy. What we discovered is that the results with the highest accuracy cannot be trusted as they can capture dependencies in the data which do not exist. 

The remaining of the paper is structured as follows Section \ref{sec:context} presents the context of our work -- CEA and thrombectomy. Section \ref{sec:results} presents the results of comparison between these two groups of patients and a discussion around the results. Finally, Section \ref{sec:conclusions} presents conclusions. 

\section{Context -- CEA, Thrombectomy, and Collected Signals}
\label{sec:context}
\noindent
Stroke is a brain injury due to inadequate blood flow to the affected region, called cerebral ischemia. The most common cause of stroke is a blood clot in a brain artery. These clots can originate from plaques in larger arteries, such as the internal carotid artery (ICA). The ICA supplies blood to the anterior part of the brain. If a plaque is detected and identified as the cause of stroke, it has to be surgically removed. This procedure is called carotid endarterectomy. During CEA, the surgeon needs to cross-clamp the ICA in order to be able to remove the plaque. Clamping the ICA cuts the blood flow to that side of the brain and causes reduced blood flow to the brain cells to various degrees depending on blood flow from the ICA of the opposite side. Thus, ischemia can occur.

A stroke can be treated during the acute phase. A novel and effective treatment is endovascular thrombectomy.\cite{powers20152015} Thrombectomy is the mechanical removal of the blood clot from the affected artery. The earlier trombectomy is performed, the more brain cells are salvaged, and the better the outcome for the patient. Time is brain. Onset of thrombectomy can be as fast as minutes after diagnosis, making it challenging to connect the desired monitoring for research. The duration of the procedure can vary between minutes and over an hour.

We collected high frequency physiological time series data from several signals, of which the most interesting are described below. The monitored period is divided in phases or events and labelled accordingly. The labels were pre-defined with regards to the depth of anaesthesia, which by affecting the blood circulation and brain activity, is expected to alter the input signals differently. Labels of interest are “clamp”, which refers to clamping of the ICA during CEA, and “Flow”, which refers to the time after a blood clot have been removed during a thrombectomy resulting in restoration of blood flow.

The following signals and features were used in the analysis:
\begin{itemize}
    \item $Electrocardiogram (ECG)$ – A recording of the electrical activity of the heart. Features extracted are inter-beat-interval, which is inversely related to heart rate, and several time domain heart rate variability (HRV) features. HRV is, independent of the heart rate, a biomarker reflecting activity in the central and autonomous nervous systems.\cite{shaffer2017overview} HRV has been shown to be affected in several diseases, such as strokes and myocardial infarctions, and is associated with increased mortality. Our hypothesis is that HRV changes occur fast after cerebral ischemia and could thus be used for its detection.
    \item $Electroencephalogram (EEG)$ – EEG is a recording of the electric activity of the superficial part of the brain, the cerebral cortex. The signal is quantified in the frequency domain resulting in 4 variables for each channel. CEA patients are monitored using 6 channels and thrombectomy patients using 2 channels. Cerebral ischemia can be detected on EEG \cite{foreman2012quantitative}, but sensitivity and specificity are not optimal. EEG can be used to detect ischemia during CEA, and it is considered standard of care in some centres.
    \item $Near-infrared spectrometry (NIRS)$ – Measures the oxygen saturation of the superficial parts of the frontal brain.
    \item Other variables include arterial blood pressure (ABP), oxygen saturation in blood (SpO2), and respiratory rate.
\end{itemize}

During a monitoring session, we use the Moberg CNS monitor to collect the data. This device is capable of taking in data from all other patient monitoring devices in a time-synchronized fashion. After a monitoring session is finished, the data is exported from the Moberg monitor and cleaned from artefacts manually on a PC. The cleaned data is then exported in matlab-format and processed for machine learning analysis using Python. 

We quantified all the data per minute and labelled each minute (data point) according to the phase of the surgery. This provided us with a data set that was used to train a machine learning model. We then evaluated the performance of different machine learning models in classifying the labels.

\section{Results and Discussion}
\label{sec:results}
\noindent

\begin{table}[!tb]
\scriptsize
\begin{tabular}{@{}llllllll@{}}
\rowcolor[HTML]{656565} 
{\color[HTML]{FFFFFF} \textbf{Subj.}} &
  {\color[HTML]{FFFFFF} \textbf{\begin{tabular}[c]{@{}l@{}}Data\\ points\end{tabular}}} &
  {\color[HTML]{FFFFFF} \textbf{\begin{tabular}[c]{@{}l@{}}UE\end{tabular}}} &
  {\color[HTML]{FFFFFF} \textbf{\begin{tabular}[c]{@{}l@{}}AB\\ acc.\end{tabular}}} &
  {\color[HTML]{FFFFFF} \textbf{\begin{tabular}[c]{@{}l@{}}CART\\ acc.\end{tabular}}} &
  {\color[HTML]{FFFFFF} \textbf{\begin{tabular}[c]{@{}l@{}}Random\\ forrest\\ acc.\end{tabular}}} &
  {\color[HTML]{FFFFFF} \textbf{\begin{tabular}[c]{@{}l@{}}SVM\\ acc.\end{tabular}}} &
  {\color[HTML]{FFFFFF} \textbf{\begin{tabular}[c]{@{}l@{}}Logistic\\ regression\\ acc.\end{tabular}}} \\
C001 & 132 & 6 & 0.78 & 0.78 & 0.93 & 0.74 & 0.78 \\
\rowcolor[HTML]{C0C0C0} 
C004 & 105 & 7 & 0.76 & 0.67 & 0.9  & 0.71 & 0.71 \\
C005 & 86  & 6 & 0.61 & 0.72 & 0.61 & 0.44 & 0.72 \\
\rowcolor[HTML]{C0C0C0} 
C007 & 92  & 7 & 0.84 & 0.84 & 0.84 & 0.58 & 0.63 \\
C008 & 109 & 7 & 0.64 & 0.59 & 0.86 & 0.77 & 0.73 \\
\rowcolor[HTML]{C0C0C0} 
C009 & 149 & 6 & 0.83 & 0.9  & 0.93 & 0.87 & 0.6  \\
C010 & 68  & 4 & 0.93 & 0.93 & 0.93 & 0.79 & 0.36 \\
\rowcolor[HTML]{C0C0C0} 
T001 & 5   & 2 & 1.0  & 1.0  & 0.0  & 0.0  & 0.0  \\
T002 & 32  & 3 & 1.0  & 1.0  & 1.0  & 1.0  & 1.0  \\
\rowcolor[HTML]{C0C0C0} 
T003 & 24  & 2 & 0.6  & 0.8  & 1.0  & 0.8  & 0.8  \\
T005 & 18  & 2 & 0.25 & 0.25 & 0.5  & 0.25 & 0.75 \\
\rowcolor[HTML]{C0C0C0} 
T006 & 30  & 2 & 0.83 & 0.83 & 0.83 & 0.83 & 0.83 \\
T010 & 18  & 2 & 0.75 & 0.75 & 1.0  & 0.0  & 1.0 
\end{tabular}
\label{table1}
\caption{Results of the machine learning analysis for each patient. UE -- Unique Events; AB -- AdaBoost; acc. -- accuracy.}
\end{table}

\begin{figure}[!htb]
    \centering
    \includegraphics[width=\columnwidth]{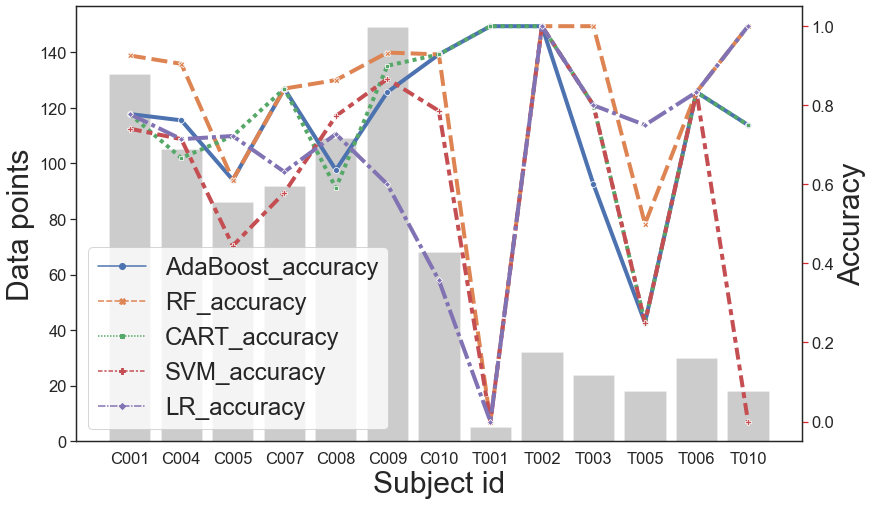}
    \caption{Bar plot showing the amount of data points in each patients data set, overlayed of line plots showing accuracy results of each algorithm for each patient.}
    \label{fig:accuracy_plot}
\end{figure}

After including 10 patients to each group, 7 CEA patients and 6 thrombectomy patients were eligible for further analysis. In six out of seven exclusions, the reason was related to inadequate quality of one or several signals. 
Results are presented in Table 1 and illustrated in Figure 1.

There are clear differences between results of CEA patients and thrombectomy patients. First, the data sets of the CEA patients were larger than those of the thrombectomy patients. This is a result of CEA being a several hour long procedure, whereas thrombectomies are usually done within an hour.
Another cause for this discrepancy in the number of data points, is the quality of data recorded. CEA is a planned procedure. This means that there is time to prepare The patient, the equipment and the OR. Quality control of all the signals is done after all the monitors are connected to the patient, and any signal that is sub-optimal can usually be corrected.
Thrombectomies are urgent procedures where the patient is constantly surrounded by medical personnel delivering care. This poses a great challenge on doing research. No intervention can be allowed to delay a thrombectomy. Therefore, attaching the EEG and NIRS-electrodes have to be done parallel to all other work, and without being in the way. We manage to do that in under 60 seconds. Yet the issue is that quality of the signal can be controlled first a few minutes later, and by that time, the patients head will be under the X-ray field and not approachable. So electrodes and wires cannot be adjusted until the procedure is finished, leaving us with signals of sub-optimal quality for the majority of the procedure. This time period with artefacts will be removed during data cleaning resulting in small data sets.

There is also a clear difference in the performance of the algorithms between CEA and thrombectomy patients. The algorithms performance is more consistent for CEA patients in comparison with thrombectomy patients, both between patients and between algorithms for each patient. To understand what causes these differences, we will discuss one patient from each group in detail.
For CEA patient C009, the total recorded time was 430 minutes. After data cleaning, the data set for this patient included 149 data points (minutes). This means that even where we have a controlled environment for data collection, the majority of data was excluded. The reason for so much data loss is artefacts in the EEG-signal caused by electrocautery, a surgical tool that uses electric current to burn tissue. Unfortunately, there is no way around this problem. 
Yet, even if a large amount of data is excluded, there still remains a large data set that can be used to train and test different ML-algorithms.  As we can see in Table 1, the results are satisfactory and consistent between different algorithms. 

\begin{figure}
    \centering
    \includegraphics[width=\columnwidth]{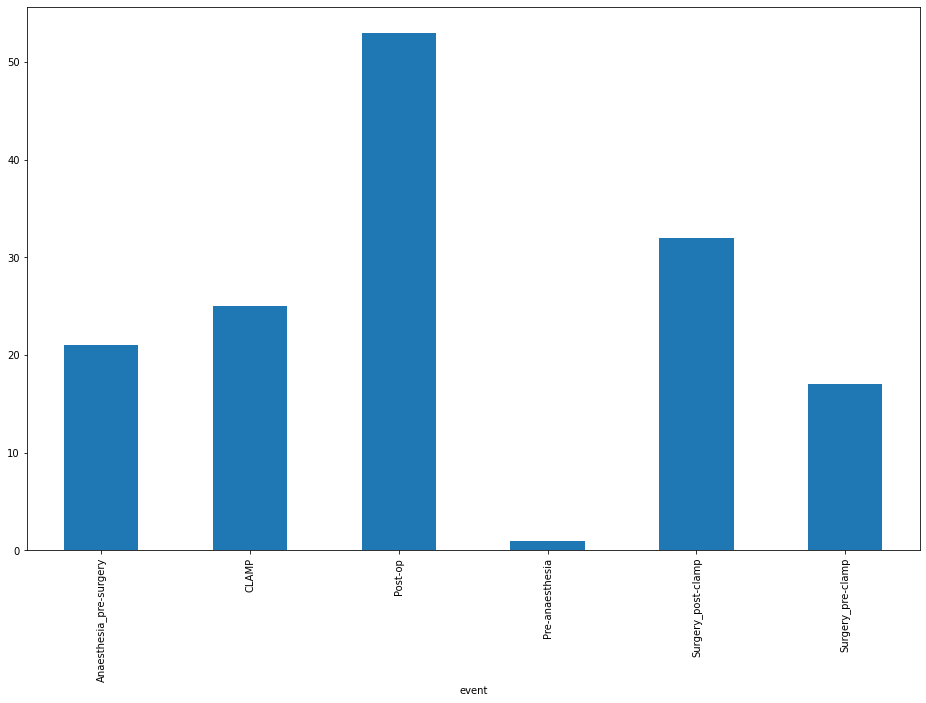}
    \caption{Number of data points for each label in the data set of patient C009}
    \label{fig:event9}
\end{figure}

\begin{figure}
    \centering
    \includegraphics[width=\columnwidth]{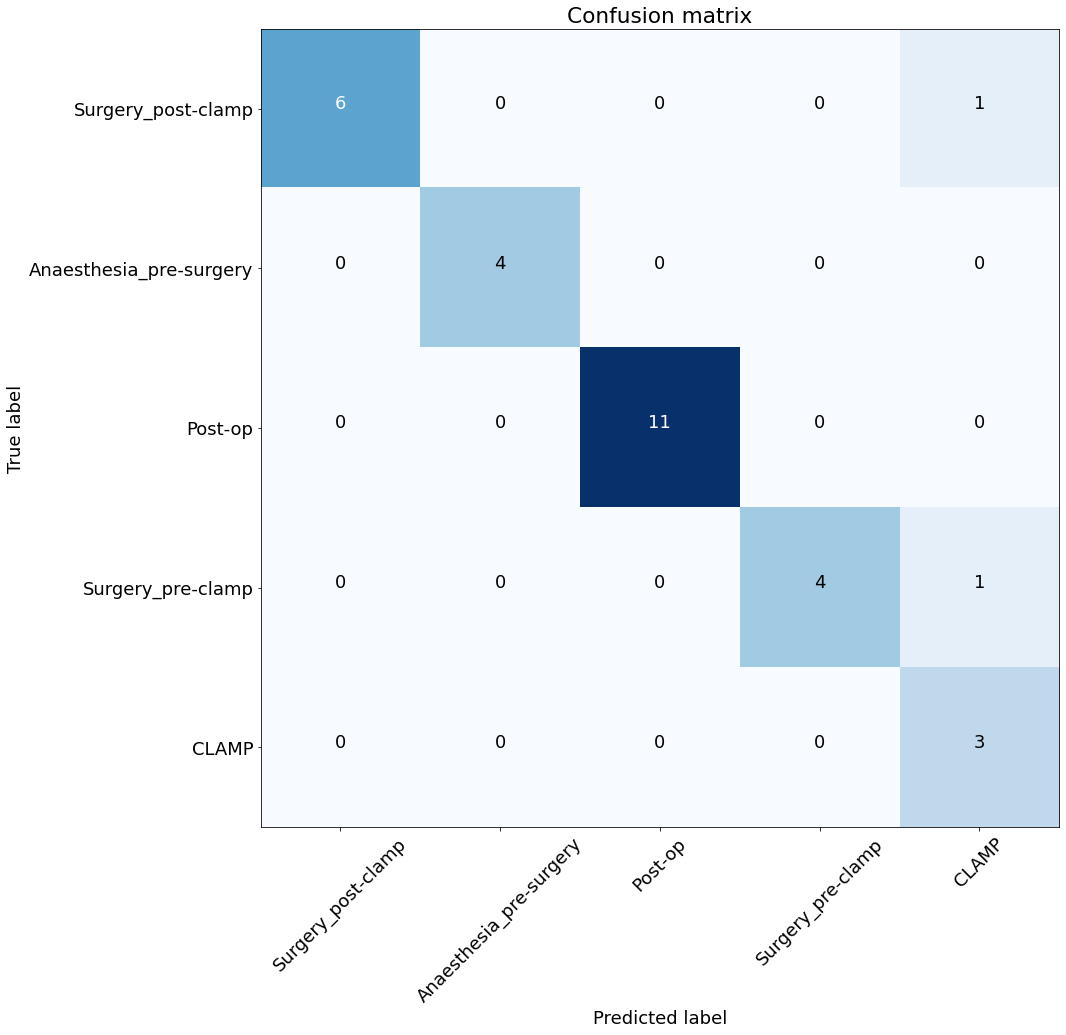}
    \caption{Confusion matrix of the results of running a random forest model on C009 data set. }
    \label{fig:con9}
\end{figure}

Since CEA is a much longer procedure than thrombectomy, the timeline can be divided into more phases or events. For C009, there were 6 unique events, or labels - see Figure 2. This should make the results more trustworthy, since the events then can have true unique profiles.  
For example, if you compare the event \texttt{CLAMP}, with \texttt{anaesthesia\_pre-surgery}, where the depth of anaesthesia is different, or post-op, where the patient is awake, one might expect an algorithm to be able to differentiate between these events, since there are so many signals that would clearly be different. So much that even a human anaesthesiologist would be able to make this differentiation.
If one on the other hand compares the events \texttt{surgery\_pre-clamp}, \texttt{CLAMP}, and \texttt{surgery\_post-clamp}, the only difference between these events is the clamping of the ICA.  Therefore, it is more impressive when an algorithm can make this distinction - see Figure 3.

For thrombectomy patient T002, the total recording time was 117 minutes. After data cleaning, 32 remained for analysis. There are 3 unique events - see Figure 4, \texttt{Anaesthesia\_NoFlow}, where the patient is under general anaesthesia before the removal of the blood clot, \texttt{Anaesthesia\_Flow}, where the patient is still under general anaesthesia but the blood clot has been removed and blood flow to the affected part of the brain restored, and \texttt{Awake\_Flow}, which stands for the period of observation after the patient have been awakened from anaesthesia.
 
In theory, the thrombectomy timeline could also be divided in more types of events, but because of the few available data points in total, it would be impractical. In the case of T002 for example, there are only 7 data points labelled \texttt{Anaesthesia\_NoFlow}. Apart from being the event with ongoing cerebral ischemia, it is also where anaesthesia and therapy to control blood pressure is started, before reaching a steady state, meaning that this can be expected to be different from the next event simply because it occurs first chronologically.

It is indeed interesting that the algorithms frequently gives accuracy values in the extremes for thrombectomy patients. For patient T002, all the models had an accuracy of 1. These results can clearly not be trusted because they are due to overfitting. The overfitting  problem in this case is a result of a small data set.

It is unclear if there is a feasible way around the problem of small data sets in the thrombectomy cases. One idea is to increase the amount of data points by quantifying the data per thirty seconds instead of one minute. This doubles the amount of data points. It would still, however, the confounding and bias problems discussed above.

To make the results more trustworthy, we need larger data sets from longer recordings, especially prior to the removal of the clot. Using our current set-up, this would practically be impossible due to the unplanned nature and urgency of the procedure. Starting simpler monitoring with ECG in the ambulance might be possible and would gain longer recordings (up to 1 hour). This approach poses logistic challenges and requires resources which may not be justified to answer this particular research question.

Another possible solution is to exclude one or more signals, as the majority of the excluded data is due to artefacts in the EEG. Excluding EEG would even give longer recordings, but with information loss -- the signal cannot be derived from other ones. EEG is the signal that is by far the most sensitive to disturbances and as EEG takes time to attach to a patients head and is not mobile, it can only be done once the patient is on the operating table. ECG on the other hand can be recorded using a small battery driven device that takes only seconds to attach. It yields a stable signal that is resistant to disturbances, increasing the amount of data in the data set.

\begin{figure}
    \centering
    \includegraphics[width=\columnwidth]{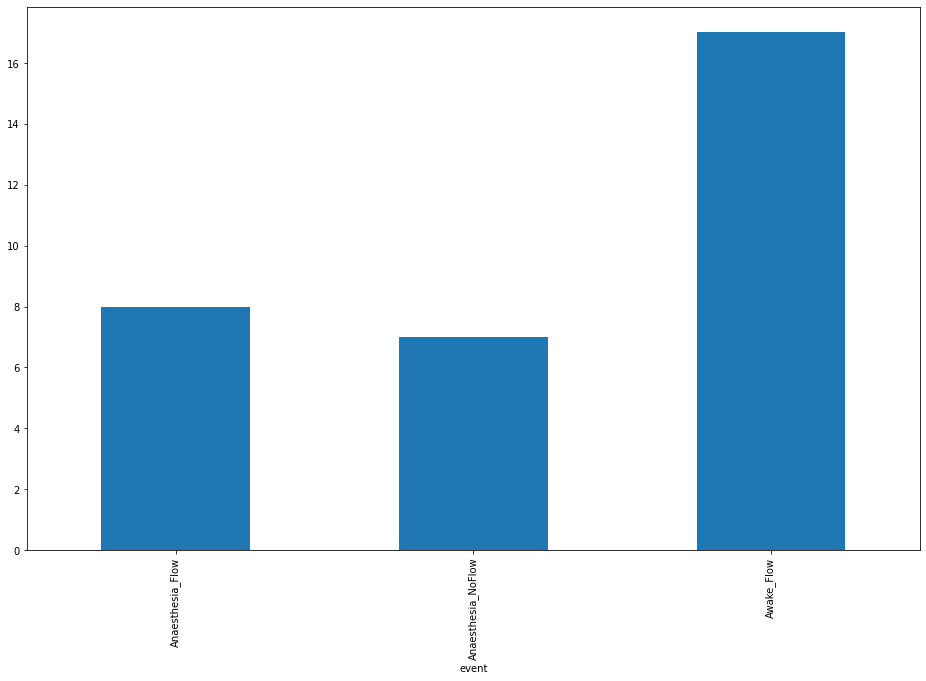}
    \caption{Number of data points for each label in the data set of patient T002}
    \label{fig:event2}
\end{figure}

\section{Conclusions}
\label{sec:conclusions}
\noindent
Using the same setup of equipment and pipeline for data collection, cleaning, and analysis yields different results when comparing CEA and thrombectomy settings. The difference in duration and urgency of each procedure has major effects on the results. What is even more interesting, is that the high accuracy of some of the analyses is not justified by the number of data points or the procedure, but can be caused by noise. This means that we need to complement these analyses with deeper scrutiny -- especially in the areas where the intervention (thrombectomy) cannot be structured around the data collection. 

For CEA, where the procedure is planned and there is time for preparation, and where the procedure takes several hour, a large data set is produced. The machine learning models performs well on these data sets.

The acute setting in which a thrombectomy is performed, causes time deficiency and signals with bad quality, which in turn produces small data sets. The results of the machine learning models on these data sets cannot be trusted.

\bibliography{aaai}

\end{document}